\documentclass{article}

\usepackage[margin=1in]{geometry}
\usepackage{natbib}

\usepackage[utf8]{inputenc}
\usepackage[T1]{fontenc}
\usepackage{amsmath,amssymb}
\usepackage{algorithm}
\usepackage{algpseudocode}
\usepackage{booktabs}
\usepackage{subcaption}
\usepackage{xcolor}
\usepackage{pgfplots}
\pgfplotsset{compat=1.17}
\usetikzlibrary{patterns}

\usepackage{url}
\usepackage{hyperref}

\definecolor{cLink}{HTML}{1A5276}   

\definecolor{cWeb}{HTML}{8C8C8C}      
\definecolor{cAgent}{HTML}{4C72B0}    
\definecolor{cAutofyn}{HTML}{DD8452}  
\definecolor{cGrid}{HTML}{DADCE0}
\definecolor{cMedGold}{HTML}{B8912F}
\definecolor{cMedSilver}{HTML}{8E9BA6}
\definecolor{cMedBronze}{HTML}{9C6B30}
\hypersetup{
  colorlinks=true,
  linkcolor=cLink,
  citecolor=cLink,
  urlcolor=cLink,
  filecolor=cLink,
  linktoc=all,
}

\title{AutoFyn Technical Report: Non-Parametric Expert Iteration for Long-Horizon Agents}

\author{
  Adib Hasan\thanks{SignalPilot Labs} \and
  Daniel Schaffield\footnotemark[1] \and
  Akashnil Dutta\thanks{Prentis AI} \and
  Tarik Adnan Moon\footnotemark[1]
}

\begin{document}

\maketitle
\begin{abstract}
We introduce AutoFyn, an agent harness inspired by the Expert Iteration algorithm, adapting a frozen model across many rounds by updating persistent state from verified reward signals rather than model weights. Each round begins from a fresh model session, and durable information is reintroduced only through explicit interfaces such as persistent memory files, reports, and repository state. Within a round, an orchestrator explores, plans and builds many alternative approaches with specialized agents, while a task-grounded verifier verifies the work and supplies an objective reward for measuring progress. This reward is distilled back into the persistent state, which updates the effective policy for the next round. In this technical report, we formalize this loop and describe its persistent state and verification interfaces. We then demonstrate its use in three domains, namely olympiad mathematics, data science, and cybersecurity. On the six fresh problems of the 2026 International Mathematical Olympiad, every model with room to improve scores higher under AutoFyn than in its provider's own coding agent. AutoFyn also built the top-ranked agent on the Spider 2.0 dbt benchmark, and has produced $16$ maintainer-confirmed vulnerability advisories in Next.js, MetaMask, pnpm, Warp, LiteLLM, Langflow, and Open WebUI.\\\textbf{Code:} \url{https://github.com/SignalPilot-Labs/autofyn}

\end{abstract}

\section{Introduction}
\label{sec:introduction}
In 2026, it has become normal for agentic systems to operate for many hours, and even days, to complete a complex task across hundreds of model calls and tool interactions. For example, coding agents resolve real world software issues that require coordinated edits across an entire repository \citep{jimenez2024swebench, yang2024sweagent}, and iterative schemes such as reflection revise their own outputs over repeated attempts \citep{shinn2023reflexion, madaan2023selfrefine}. As the operational horizon grows, two failure modes recur in agentic systems. First, the context of a single agent accumulates until the model attends to it unevenly and its effective use degrades \citep{liu2024lost}. Second, when an agent evaluates its own work, an incorrect conclusion may pass that evaluation and condition later steps, since language models are unreliable at correcting their own reasoning without external feedback \citep{huang2024selfcorrect}, so errors accumulate across the run.

AutoFyn is designed around these two failure modes. To bound the context, AutoFyn uses an orchestrator worker model where a single orchestrator delegates work to subagents and records each round's progress to disk, then begins the next round in a fresh context initialized only from that record. To prevent errors from accumulating, the signal that closes each round is an external verification rather than the agent's own assessment. A round is treated as progress only when that signal confirms it. AutoFyn applies both together, so that what a round carries forward is bounded in size and confirmed outside the agent.

The resulting loop is analogous to Expert Iteration \citep{anthony2017thinking}, developed in \autoref{sec:method}. The correspondence, however, is structural rather than algorithmic, since AutoFyn is non-parametric and an adapted policy is not automatically a better one. This places AutoFyn alongside methods that pair a frozen model with an updated experience memory \citep{zhang2023rememberer}.

\begin{figure}[tb]
\centering
\begin{subfigure}{\linewidth}
\centering
\begin{tikzpicture}
\begin{axis}[
    ybar,
    width=0.87\linewidth,
    height=6.3cm,
    bar width=10pt,
    ymin=0, ymax=45,
    ytick={0,7,14,21,28,35,42},
    ylabel={Score (out of 42)},
    symbolic x coords={GPT-5.6 Sol, Claude Fable 5, Claude Opus 4.8, Claude Sonnet 5, GLM 5.2},
    xtick={GPT-5.6 Sol, Claude Fable 5, Claude Opus 4.8, Claude Sonnet 5, GLM 5.2},
    x tick label style={font=\small, align=center},
    enlarge x limits=0.12,
    legend style={at={(0.5,-0.16)}, anchor=north, legend columns=2, draw=none, font=\small, legend cell align=left, /tikz/every even column/.append style={column sep=10pt}},
    legend image code/.code={\draw[#1] (0cm,-0.08cm) rectangle (0.26cm,0.12cm);},
    tick label style={font=\small},
    label style={font=\small},
    axis lines*=left,
    ymajorgrids=true,
    grid style={cGrid},
    clip=false,
    every axis plot/.append style={draw=none},
]
\addplot[bar shift=-12pt, fill=cWeb] coordinates {(GPT-5.6 Sol,41) (Claude Sonnet 5,14) (GLM 5.2,14)};
\addlegendentry{Web app}
\addplot[bar shift=-12pt, pattern=north east lines, pattern color=cWeb, draw=cWeb] coordinates {(Claude Opus 4.8,14) (Claude Fable 5,28)};
\addlegendentry{Web app, no output on some problems}
\addplot[bar shift=0pt, fill=cAgent, fill opacity=0.55] coordinates {(Claude Opus 4.8,28.0) (GPT-5.6 Sol,40) (Claude Fable 5,42) (Claude Sonnet 5,23.3) (GLM 5.2,20.7)};
\addlegendentry{Provider harness (e.g.\ Claude Code)}
\addplot[bar shift=12pt, fill=cAutofyn, fill opacity=0.55] coordinates {(Claude Opus 4.8,34.7) (GPT-5.6 Sol,42) (Claude Fable 5,42) (Claude Sonnet 5,29.7) (GLM 5.2,32.0)};
\addlegendentry{AutoFyn}
\addplot[only marks, mark=*, mark size=2.4pt, draw=black!70, fill=black!55, xshift=-12pt, forget plot] coordinates {(GPT-5.6 Sol,41) (Claude Sonnet 5,14) (GLM 5.2,14) (Claude Opus 4.8,14) (Claude Fable 5,28)};
\addplot[only marks, mark=*, mark size=2.4pt, draw=cAgent!60!black, fill=cAgent!80!black, xshift=0pt, forget plot] coordinates {(GPT-5.6 Sol,40) (Claude Fable 5,42) (Claude Sonnet 5,28)};
\addplot[only marks, mark=*, mark size=2.4pt, draw=cAgent!60!black, fill=cAgent!80!black, xshift=0pt, forget plot] coordinates {(Claude Opus 4.8,21) (Claude Opus 4.8,35) (Claude Opus 4.8,28) (GLM 5.2,20) (GLM 5.2,28) (GLM 5.2,14)};
\addplot[only marks, mark=*, mark size=2.4pt, draw=cAgent!60!black, fill=cAgent!80!black, xshift=-2pt, forget plot] coordinates {(Claude Sonnet 5,21)};
\addplot[only marks, mark=*, mark size=2.4pt, draw=cAgent!60!black, fill=cAgent!80!black, xshift=2pt, forget plot] coordinates {(Claude Sonnet 5,21)};
\addplot[only marks, mark=*, mark size=2.4pt, draw=cAutofyn!65!black, fill=cAutofyn!85!black, xshift=10pt, forget plot] coordinates {(Claude Opus 4.8,35)};
\addplot[only marks, mark=*, mark size=2.4pt, draw=cAutofyn!65!black, fill=cAutofyn!85!black, xshift=14pt, forget plot] coordinates {(Claude Opus 4.8,35)};
\addplot[only marks, mark=*, mark size=2.4pt, draw=cAutofyn!65!black, fill=cAutofyn!85!black, xshift=10pt, forget plot] coordinates {(Claude Sonnet 5,26)};
\addplot[only marks, mark=*, mark size=2.4pt, draw=cAutofyn!65!black, fill=cAutofyn!85!black, xshift=10pt, forget plot] coordinates {(GLM 5.2,34)};
\addplot[only marks, mark=*, mark size=2.4pt, draw=cAutofyn!65!black, fill=cAutofyn!85!black, xshift=14pt, forget plot] coordinates {(GLM 5.2,34)};
\addplot[only marks, mark=*, mark size=2.4pt, draw=cAutofyn!65!black, fill=cAutofyn!85!black, xshift=12pt, forget plot] coordinates {(Claude Sonnet 5,35) (Claude Sonnet 5,28) (GPT-5.6 Sol,42) (Claude Fable 5,42) (Claude Opus 4.8,34) (GLM 5.2,28)};
\draw[cMedBronze, dash pattern=on 3.5pt off 2.5pt, line width=0.9pt]
  (rel axis cs:0,0.35556) -- (rel axis cs:1,0.35556)
  node[anchor=west, xshift=2pt, font=\scriptsize, text=cMedBronze] {Bronze (16)};
\draw[cMedSilver, dash pattern=on 3.5pt off 2.5pt, line width=0.9pt]
  (rel axis cs:0,0.51111) -- (rel axis cs:1,0.51111)
  node[anchor=west, xshift=2pt, font=\scriptsize, text=cMedSilver] {Silver (23)};
\draw[cMedGold, dash pattern=on 3.5pt off 2.5pt, line width=0.9pt]
  (rel axis cs:0,0.64444) -- (rel axis cs:1,0.64444)
  node[anchor=west, xshift=2pt, font=\scriptsize, text=cMedGold] {Gold (29)};
\end{axis}
\end{tikzpicture}
\caption{IMO 2026, audited score out of $42$ for each model under each harness. Bars are cell means, circles are individual runs, and the dashed lines mark the 2026 medal cutoffs.}
\label{fig:imo-scores}
\end{subfigure}

\vspace{1.2em}

\begin{subfigure}{\linewidth}
\centering
\begin{tikzpicture}
\definecolor{barother}{HTML}{9DB4C0}
\definecolor{barours}{HTML}{B23A48}
\begin{axis}[
  ybar,
  bar shift=0pt,
  width=0.86\linewidth,
  height=5.4cm,
  ymin=0, ymax=78,
  ytick={0,20,40,60},
  ylabel={Spider 2.0 dbt score},
  bar width=24pt,
  symbolic x coords={USTC-KCIL, Spider-Agent-Extended, Shadowfax-DBT, Databao, SignalPilot},
  xtick={USTC-KCIL, Spider-Agent-Extended, Shadowfax-DBT, Databao, SignalPilot},
  x tick label style={font=\footnotesize, rotate=28, anchor=east, xshift=2pt, yshift=-2pt},
  enlarge x limits=0.13,
  axis lines*=left,
  axis line style={-, draw=black!50},
  ymajorgrids=true,
  grid style={draw=black!10, line width=0.4pt},
  tickwidth=0pt,
  nodes near coords,
  point meta=explicit symbolic,
  every node near coord/.append style={font=\footnotesize, text=black, yshift=1pt},
  tick label style={font=\footnotesize},
  label style={font=\small},
  every axis plot/.append style={draw=none},
  clip=false,
]
\addplot[fill=barother] coordinates {
  (USTC-KCIL,39.71)[39.71] (Spider-Agent-Extended,39.71)[39.71]
  (Shadowfax-DBT,41.18)[41.18] (Databao,60.29)[60.29]
};
\addplot[fill=barours] coordinates {(SignalPilot,65.60)[65.60]};
\end{axis}
\end{tikzpicture}
\caption{Spider 2.0 dbt, top five entries on the public leaderboard. The SignalPilot Agent, highlighted, was built and optimized autonomously by AutoFyn. Scores are as listed on 31 July 2026.}
\label{fig:spider}
\end{subfigure}
\caption{AutoFyn performance in olympiad mathematics and data science. The olympiad scores in (a) are the audited grades of our companion study \citep{hasan2026imo}.}
\label{fig:performance}
\end{figure}
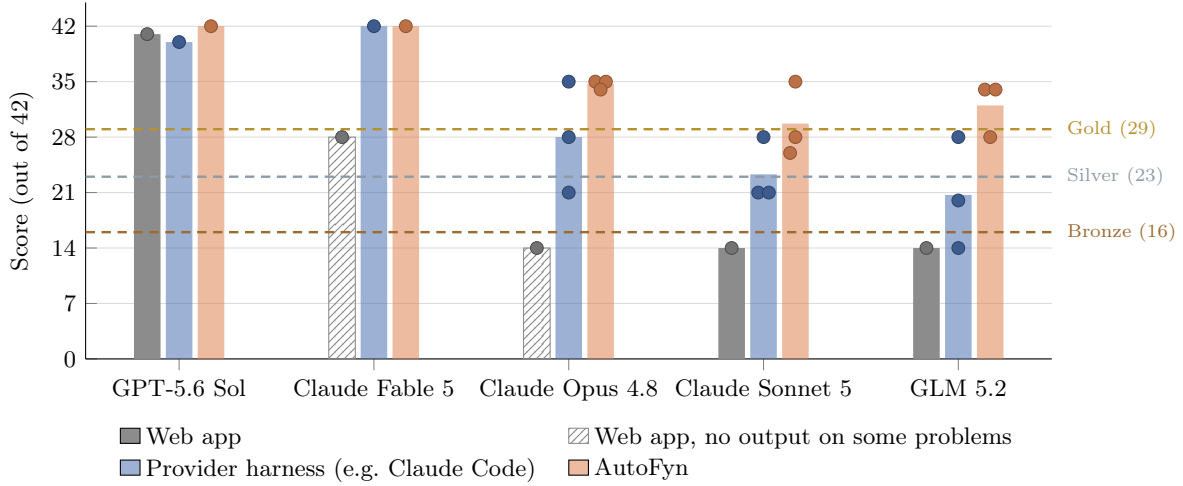
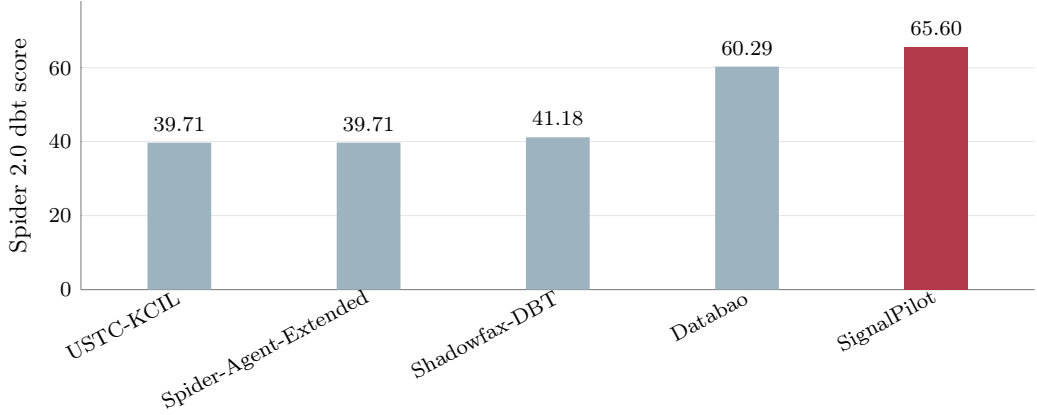

In this report, we describe AutoFyn and its performance across three domains, namely, olympiad mathematics, data science, and cyber security. On the six problems of the 2026 International Mathematical Olympiad, every model with room to improve scores higher under AutoFyn than in its provider's own coding agent (\autoref{fig:imo-scores}), and each run's process trail is fully archived. On the Spider 2.0 dbt benchmark for data science tasks, an agent AutoFyn built and optimized without human intervention tops the public leaderboard (\autoref{fig:spider}). In security auditing, the loop has found over $150$ individual vulnerabilities in widely used open source projects, of which we have submitted $43$ advisories with $16$ confirmed by their maintainers to date.

Our contributions are the following.
\begin{itemize}
\item AutoFyn as a system, a long horizon agent harness that composes context reset across rounds, search over candidate policies, an externally verified reward gate, and a distilled persistent state into a single loop.
\item A formulation of this loop as a non parametric expert iteration, in which search proposes candidates, the external verifier is the expert criterion, and distillation into persistent state is the policy update.
\item Results across three domains, an audited study on the 2026 International Mathematical Olympiad, a top ranked data science agent on the Spider 2.0 dbt benchmark, and maintainer confirmed vulnerability advisories in widely used open source projects.
\end{itemize}

\section{Related Work}
\label{sec:related}
\paragraph{Expert iteration.}
AutoFyn builds on expert iteration \citep{anthony2017thinking}, which alternates a search stronger than the current policy with an update that folds the search results back in. Self play systems realize this pattern with tree search as the expert \citep{silver2018general}, and recent work brings tree search to language model reasoning \citep{zhou2024lats}. AutoFyn instead plays the expert with a verifier grounded loop of review and revision, and its update adapts non parametric context state rather than model parameters or a search tree. Methods that pair a frozen model with an updated experience memory \citep{zhang2023rememberer} share this weightless adaptation, and reflection revises its own outputs without external grounding \citep{shinn2023reflexion, madaan2023selfrefine}. We are not aware of prior work that frames such a loop as expert iteration, and \autoref{sec:expert-iteration} states the correspondence and the point at which it stops holding.

\paragraph{Long context agents.}
It is known that long contexts are used unevenly \citep{liu2024lost} and researchers have studied ways to manage them. Proposed mechanisms include paged memory \citep{packer2023memgpt}, hierarchical working memory \citep{hu2024hiagent}, explicit context management for coding agents \citep{liu2025context}, and history folding for long running web and software tasks \citep{sun2025contextfolding, ye2025agentfold}. Closest to our own is the Ralph loop \citep{huntley2025ralph}, which reinvokes a coding agent on the same prompt indefinitely, so each iteration starts an empty context and state survives only through the repository, a plan file, and a specification directory. The shell loop itself reads no exit status, and continuation rests on human judgment of the repository. AutoFyn scores each round with an external verifier, and the orchestrator carries an artifact forward only when that score improves on the best so far (\autoref{eq:incumbent}). Fresh contexts, summarization, and bounded memory each predate our work, and AutoFyn contributes their composition with hard round boundaries, provenance bearing state, and grounded acceptance.

\paragraph{Verification.}
Verification and refinement pipelines have reported strong olympiad results by iterating a model against structured checks \citep{huang2025imogold}. AutoFyn shares this verifier driven shape and differs in its explicit persistent state and its typed verification records. What a check establishes also depends on the harness around it, since the design of the agent computer interface shapes agent behavior in software engineering \citep{yang2024sweagent}. Recent work measures the harness and the model together rather than the model alone, benchmarking harness configurations across model backends \citep{yao2026harnessbench} and transferring a single optimized harness across five held out models on olympiad level problems \citep{lee2026metaharness}.

\section{Method}
\label{sec:method}
\subsection{Problem Setting}
\label{sec:setting}

Let $G$ be the goal, $E$ the environment, and $B$ the budget expressed as wall clock time and a number of rounds. Let $\mathcal{R}$ be the set of roles and $\Theta = \{\theta_r : r \in \mathcal{R}\}$ the assignment of a base language model to each role. The parameters in $\Theta$ stay frozen for the entire episode, so roles may resolve to different model tiers but no assignment changes during a run. Where one model serves every role, $\Theta$ reduces to a single frozen $\pi_\theta$. Let $M_t$ be the persistent state at round $t$, written to disk and reconstructed into a fresh context each round. Let $V$ be the verifier the environment exposes, which maps a candidate artifact $a$ to a typed record
\begin{equation*}
y = V(a; E) = (s, e, \ell, v),
\end{equation*}
with status or score $s$, evidence bundle $e$, assurance class $\ell$, and verifier identity $v$. The assurance class orders evidence by strength, from model self assessment through separate context review, executable checks, and independent human grading, up to a machine checked certificate. A later round reads the record rather than a bare accept or reject, so it acts on the reason a candidate failed.

A round spends part of $B$ and produces one or more candidate artifacts together with a distilled update to the state. Five procedures carry it. The context construction $K$ assembles a session from the goal, the state, and an interaction history, the orchestrated search $\mathcal{S}$ returns a trajectory $\tau_t$ and a candidate set $\mathcal{A}_t$, the selection $A$ chooses among those candidates on their records, the retention $S_{\mathrm{inc}}$ updates the incumbent across rounds, and the distillation $U$ folds the round into the state. We write $\operatorname{accept}(y) = 1$ when the record $y$ meets the acceptance condition the environment defines. The episode ends under the control policy of \autoref{sec:stopping}, and the reported output is the incumbent artifact retained under the acceptance rule of \autoref{sec:selection}.

\subsection{Core Design Principles}
\label{sec:principles}

\paragraph{Fresh sessions over explicit state.}
Every round begins a fresh model session. The previous round's conversation is not reloaded, and information returns only through a fixed set of durable interfaces, namely the persistent memory files, the repository and the artifacts it references, prior reports when requested, and the record of user activity. These carry provenance with them, so an evaluation outcome arrives with the evidence behind it and a failed approach arrives marked as failed. Context size therefore tracks the state rather than the elapsed run.

\paragraph{Search over candidates.}
Rather than commit to a single line of attack, the orchestrator dispatches role specialized agents to explore alternative approaches, produce competing plans or proof strategies, build one or more candidates, and review them, routing a rejected build back to the builder and a rejected plan back to the planner. This branching over approaches and revisions is the search half of expert iteration, structured by role and by phase rather than by an explicit search tree, with its breadth in a round set by the budget.

\paragraph{Grounded acceptance.}
A round is judged by evidence from the environment rather than by the model's assessment of its own output (\autoref{sec:verification}). Self assessment may steer the search within a round, since the model must decide what to try next, but it cannot establish that the goal has been met. Language models are unreliable at judging their own reasoning without external feedback \citep{huang2024selfcorrect}, so AutoFyn does not let an agent certify its own success.

\subsection{The AutoFyn Round}
\label{sec:round}

An orchestrator drives each round. It reads the persistent state, decides the highest value step toward the goal, and routes that step through exploration, planning, an optional plan review, building, and build review, each served by a role specialized agent. It does not explore the codebase, design solutions, or write code itself beyond small fixes. Whether any two agents run concurrently depends on how the dispatch was issued, so generating and comparing alternative candidates is branching search even when the branches run in sequence.

Control of the loop lives outside the model, in a surrounding process. That process starts a fresh orchestrator session per round, reads persistent metadata, archived memory, user activity, and the previous round's report index to construct the round, and after the round records the round summary, commits and pushes the work, and decides whether to begin another round. The git save and push behavior is performed by the harness rather than by the orchestrator, and per round reports together with the memory directory are archived outside the sandbox and restored on resume. \autoref{alg:round} states the loop.

\begin{algorithm}[t]
\caption{The AutoFyn round loop}
\label{alg:round}
\begin{algorithmic}[1]
\State \textbf{input:} goal $G$, frozen role assignment $\Theta$, orchestration procedure $\Pi$, verifier $V$, budget $B$
\State initialize persistent state $M_0$ from $G$; incumbent $I_0 \gets \varnothing$
\State $t \gets 0$
\While{control policy permits continuation}
  \State $C_t \gets K(G, M_t, \varnothing)$ \Comment{fresh session; durable state only, no inherited transcript}
  \State $(\tau_t, \mathcal{A}_t) \gets \mathcal{S}(\Theta, \Pi, C_t; B_t)$ \Comment{orchestrated search; trajectory and candidates}
  \State $y_{t,i} \gets V(a_{t,i}; E)$ for each $a_{t,i} \in \mathcal{A}_t$ \Comment{evidence bearing records}
  \State $a_t \gets A(\mathcal{A}_t, \{y_{t,i}\}_i, I_t)$ \Comment{within round selection}
  \State $I_{t+1} \gets S_{\mathrm{inc}}(I_t, a_t, y_t)$ \Comment{cross round incumbent update}
  \State $M_{t+1} \gets U(M_t, \tau_t, a_t, y_t)$ \Comment{lossy distillation into typed state}
  \State archive reports and repository state; discard the live model session
  \State $t \gets t + 1$
\EndWhile
\State \textbf{return} incumbent $I_t$
\end{algorithmic}
\end{algorithm}

\subsection{Non-Parametric Expert Iteration}
\label{sec:expert-iteration}

Define the contextual policy induced at round $t$ by
\begin{equation*}
\pi^{M_t}_\Theta(x \mid h) = \Pi\big(\Theta, K(G, M_t, h)\big),
\end{equation*}
where $h$ is the within round interaction history and $\Pi$ composes the role assigned models into the orchestrated search. Each round opens with $h$ empty, so $C_t = K(G, M_t, \varnothing)$ and the state alone carries a round into the next. The parameters in $\Theta$ are fixed, yet changing $M_t$ changes the distribution over subsequent behavior, so $M_{t+1} \neq M_t$ implies $\pi^{M_{t+1}}_\Theta \neq \pi^{M_t}_\Theta$ in general. The policy thus changes across rounds with no parameter update, which is a contextual update rather than training in the usual sense.

\subsection{Comparison with Classical Expert Iteration}
\label{sec:comparison-ei}

Classical expert iteration has three parts, an apprentice policy, an expert procedure that spends extra compute to produce actions stronger than the apprentice's, and an update that folds the expert's targets back into the apprentice. AutoFyn instantiates all three within a round, as \autoref{tab:correspondence} sets out. Its expert is the review and revision loop, which grades a candidate against the environment, diagnoses the failure, and routes the revision back to planning or building until a candidate is accepted. The verifier $V$ grounds that loop rather than constituting the expert itself, and the expert spends more compute than a single apprentice sample, across several candidates and rounds of revision. The update is the distillation $U$, which folds the accepted candidate and its findings into $M_t$ rather than into the weights.

\begin{table}[t]
\centering
\caption{Structural correspondence between classical expert iteration and AutoFyn.}
\label{tab:correspondence}
\begin{tabular}{ll}
\toprule
Classical expert iteration & AutoFyn \\
\midrule
Apprentice policy $\pi_{\theta_t}$ & Contextual policy $\pi^{M_t}_\Theta$ \\
Apprentice proposes candidates & Explore, plan, and build produce candidate artifacts \\
Expert procedure & Verifier grounded review and revision \\
Grounding signal within the expert & Grounded verifier $V$ \\
Expert target & Selected accepted candidate and its findings \\
Parameter update $\theta_{t+1}$ & Persistent state update $M_{t+1}$ \\
Updated apprentice & Newly conditioned policy $\pi^{M_{t+1}}_\Theta$ \\
\bottomrule
\end{tabular}
\end{table}

The correspondence is structural rather than algorithmically identical. AutoFyn optimizes no parameters and constructs no explicit expert policy, updating instead the non parametric state that conditions later actions. It also lacks the guarantee that makes a game playing expert reliable, since a Monte Carlo tree search provably improves on its apprentice whereas AutoFyn's verifier can accept a wrong candidate or reject a right one. Improvement is therefore established by later verified performance rather than assumed from the update.

\subsection{Verification}
\label{sec:verification}

The environment produces the signal, not the model, so a check is reproducible from the artifact and the environment alone. Each one is also narrow. A passing test speaks only to its cases, a benchmark only to its evaluator, and a failed refutation only to the cases it searched, so the record names the property it checks and leaves the rest open.

In data science the check is the benchmark evaluator, which scores a submission against held out expected outputs and returns a per case breakdown. The breakdown rather than the bare score drives the next round, since a failed case list tells it where to look. In security auditing the check is executing the candidate exploit against the target build, which demonstrates a reachable path rather than the full scope or severity of the vulnerability. Regression tests pass or fail throughout and are diagnostic either way.

In olympiad mathematics the check only refutes, since it cannot establish that the logic of a proof is sound. Where the proof claims a bound, AutoFyn searches small and degenerate cases for a counterexample. Where it claims an algebraic computation, a symbolic algebra package looks for a discrepancy. A counterexample refutes the step it targets and an unsuccessful search establishes nothing, so the signal is reproducible from the artifact rather than dependent on model judgment. This suffices at olympiad level, where the claims inside a proof are concrete enough to attack by search. Research level problems would need a proof assistant such as Lean.

\subsection{Selection, Memory, and Incumbent Retention}
\label{sec:selection}

When candidates carry a comparable grounded score $q$, the retention $S_{\mathrm{inc}}$ keeps the better accepted artifact,
\begin{equation}
\label{eq:incumbent}
I_{t+1} =
\begin{cases}
a_t, & \operatorname{accept}(y_t) = 1 \ \land\ q(a_t) > q(I_t), \\
I_t, & \text{otherwise},
\end{cases}
\end{equation}
so that $q(I_{t+1}) \ge q(I_t)$ under deterministic comparable scoring. The orchestrator applies this rule when it writes the round's outcome into the persistent state, leaving $I_t$ and the recorded goal untouched on a round that improves on neither, so the rule sits in the protocol layer of \autoref{tab:enforcement} rather than the harness layer. It bounds the incumbent, not the contextual policy, which carries no such guarantee since $U$ is lossy.

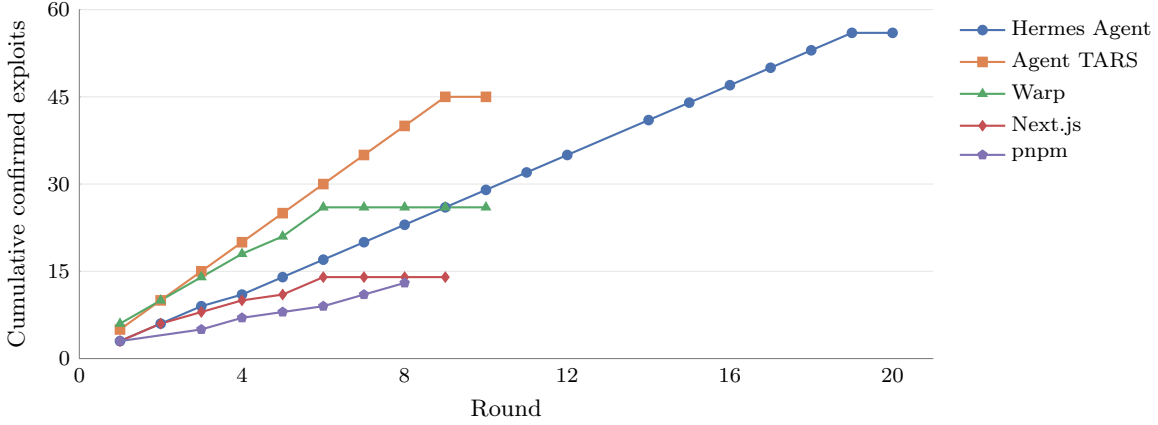
\begin{figure}[t]
\centering
\begin{tikzpicture}
\definecolor{cHermes}{HTML}{4C72B0}
\definecolor{cTars}{HTML}{DD8452}
\definecolor{cWarp}{HTML}{55A868}
\definecolor{cNext}{HTML}{C44E52}
\definecolor{cPnpm}{HTML}{8172B3}
\begin{axis}[
  width=0.78\linewidth,
  height=6.2cm,
  xlabel={Round},
  ylabel={Cumulative confirmed exploits},
  xmin=0, xmax=21,
  ymin=0, ymax=60,
  xtick={0,4,8,12,16,20},
  ytick={0,15,30,45,60},
  axis lines=left,
  axis line style={-, draw=black!50},
  ymajorgrids=true,
  grid style={draw=black!10, line width=0.4pt},
  tickwidth=0pt,
  tick label style={font=\footnotesize},
  label style={font=\small},
  legend style={
    font=\footnotesize,
    at={(1.02,1)}, anchor=north west,
    draw=none, fill=none, row sep=1pt,
  },
  legend cell align=left,
  mark size=1.6pt,
]
\addplot[color=cHermes, mark=*, thick] coordinates {
  (1,3) (2,6) (3,9) (4,11) (5,14) (6,17) (7,20) (8,23) (9,26) (10,29)
  (11,32) (12,35) (14,41) (15,44) (16,47) (17,50) (18,53) (19,56) (20,56)
};
\addlegendentry{Hermes Agent}
\addplot[color=cTars, mark=square*, thick] coordinates {
  (1,5) (2,10) (3,15) (4,20) (5,25) (6,30) (7,35) (8,40) (9,45) (10,45)
};
\addlegendentry{Agent TARS}
\addplot[color=cWarp, mark=triangle*, thick] coordinates {
  (1,6) (2,10) (3,14) (4,18) (5,21) (6,26) (7,26) (8,26) (9,26) (10,26)
};
\addlegendentry{Warp}
\addplot[color=cNext, mark=diamond*, thick] coordinates {
  (1,3) (2,6) (3,8) (4,10) (5,11) (6,14) (7,14) (8,14) (9,14)
};
\addlegendentry{Next.js}
\addplot[color=cPnpm, mark=pentagon*, thick] coordinates {
  (1,3) (3,5) (4,7) (5,8) (6,9) (7,11) (8,13)
};
\addlegendentry{pnpm}
\end{axis}
\end{tikzpicture}
\caption{Cumulative confirmed exploits by round in five security audit runs, each an independent AutoFyn episode against a different target. A round adds to the count only when a new exploit fires against the target build, so the curves are monotone by construction and flatten once a run turns to consolidating what it has. The pnpm run is truncated at round $8$, after which it spent a long stretch without a confirmed addition before closing a final chain at round $129$.}
\label{fig:security-rounds}
\end{figure}

\autoref{fig:security-rounds} shows the rule running in five security audits, where $q$ counts the exploits confirmed against the target build. Each curve rises while the search finds new reachable paths and flattens once the run turns to consolidating what it has. The Hermes Agent run reaches $56$ confirmed exploits over $19$ rounds, while the pnpm run closes thirteen by round $8$ and then spends a long stretch without a confirmed addition before a final chain at round $129$. A flat stretch is a round that produced no accepted improvement and left the incumbent in place.

The persistent state is partitioned as $M_t = (G_t, H_t, R_t, W_t, I_t)$, comprising the goal and its user amendments, the evaluation history, the distilled rules and role specific lessons, the working status of hypotheses, failures, and next steps, and the incumbent artifact with its evidence. Each subagent role owns a memory file into which it distills rules specific to that role, so a builder accumulates build knowledge and a reviewer accumulates review knowledge without either polluting the other. Rules are written as commands rather than observations, carry the round and reason they were learned, and are retired when the code they concern is gone. The working status keeps hypotheses, failures, and next steps in sections of their own, so a verified outcome stays distinguishable from speculation.

\subsection{What the Architecture Enforces}
\label{sec:enforcement}

Some of the behavior described above is enforced mechanically by the harness. The rest is requested through prompts and holds only insofar as an archived run confirms it. \autoref{tab:enforcement} sorts representative behaviors into these categories.

\begin{table}[t]
\centering
\caption{Assurance level of representative system behaviors.}
\label{tab:enforcement}
\begin{tabular}{ll}
\toprule
Behavior & Assurance \\
\midrule
Fresh model session per round & Enforced by harness \\
Round transitions, archival, resume & Enforced by harness \\
Git commit and push after a round & Enforced by harness \\
Time lock on early termination & Enforced by harness \\
Frozen role to model assignment for a run & Enforced by harness \\
Orchestrator updates \texttt{run\_state.md} each round & Required by protocol \\
Orchestrator delegates rather than codes directly & Required by protocol \\
Every build is reviewed before acceptance & Required by protocol \\
Verified facts kept distinct from hypotheses & Required by protocol \\
Concurrency of same type specialists & Observed per run \\
\bottomrule
\end{tabular}
\end{table}

\subsection{Stopping and Output Selection}
\label{sec:stopping}

Termination depends on whether the run is time locked. In an unlocked run the orchestrator may stop as soon as the goal is accepted. In a time locked run the harness denies an early end of session until the budget expires or the user intervenes, after which the loop may allow a single grace round for consolidation where configured. Episode duration is therefore not the time to reach the final candidate, which matters when reading the timings we report. The output is the incumbent retained under \autoref{eq:incumbent}, together with the evaluation history documenting how it was reached.

\section{Applications}
\label{sec:applications}
We report results in three domains, ordered by how strong a verifier the environment affords. In olympiad mathematics a proof cannot be executed and acceptance rests on refutation. In data science a benchmark evaluator settles it, and in security auditing a candidate exploit either fires against the target build or does not. \autoref{tab:applications} sets out how each was run.

\begin{table}[!ht]
\centering
\caption{How each application was run, with the verifiers as described in \autoref{sec:verification} and intervention counted over the episode alone.}
\label{tab:applications}
\small
\setlength{\tabcolsep}{5pt}
\begin{tabular}{llllll}
\toprule
Domain & Model & Budget & Verifier & Intervention & Evidence \\
\midrule
Olympiad mathematics & several        & $20$ h per run & refutation          & none               & audited grades \\
Data science         & Opus 4.5/4.8   & unlimited      & per case evaluator   & none               & leaderboard \\
Security auditing    & Opus 4.5/4.8   & unlimited      & exploit reproduction & stop a stalled run & advisory IDs \\
\bottomrule
\end{tabular}
\end{table}

\subsection{Olympiad Mathematics: IMO 2026}
\label{sec:imo}

The six problems of the 2026 International Mathematical Olympiad were released after every model's training cutoff. We ran five models on all six through three harnesses, the provider's web interface, the provider's own coding agent, and AutoFyn held fixed across every model, and graded the proofs under a completion-based standard that awards $7$ only for a complete argument and $0$ when any load-bearing step is left unproved. Each run was audited by a frontier model other than the one that wrote the proof, and a panel of three past IMO medalists reviewed those audits. All of this ran after the episodes had closed, so no grade reached the loop that produced the proof. The full matrix, the grading standard, and the analysis of where proofs fail are the subject of our companion study \citep{hasan2026imo}, which archives every graded write-up, audit report, and process trail.

\autoref{tab:imo} sets each model's AutoFyn total against the same model in its provider's own coding agent. Every model with room to improve scores higher under AutoFyn than in its provider's agent, by $2.0$ points for GPT-5.6~Sol and by $11.3$ for GLM~5.2, which moves from $20.7$ to $32.0$ and crosses a medal band with no change to the model backend. Only Claude Fable~5 has no room, completing the contest in both harnesses. The gain is bounded above by the model, since Problem~3 goes unsolved in all eighteen sub-frontier runs across both agent harnesses. We do not isolate what produces the difference. The harnesses vary in episode length, tool access, retrieval, and multi-agent structure, so these totals compare deployments rather than the loop alone.

\begin{table}[t]
\centering
\caption{IMO 2026 totals out of $42$ for each model in its provider's coding agent and in AutoFyn. The six sub-frontier cells are means over three independent runs, so those entries may be fractional. Scores are the audited grades reported in our companion study \citep{hasan2026imo}, which describes the grading standard and the audit in full.}
\label{tab:imo}
\begin{tabular}{llccc}
\toprule
Model & Provider agent & Agent total & AutoFyn total & Gain \\
\midrule
GPT-5.6 Sol      & Codex       & 40.0 & 42.0 & $+2.0$ \\
Claude Fable~5   & Claude Code & 42.0 & 42.0 & $+0.0$ \\
Claude Opus~4.8  & Claude Code & 28.0 & 34.7 & $+6.7$ \\
Claude Sonnet~5  & Claude Code & 23.3 & 29.7 & $+6.3$ \\
GLM~5.2          & Zcode       & 20.7 & 32.0 & $+11.3$ \\
\bottomrule
\end{tabular}
\end{table}

\subsection{Data Science: Spider 2.0 dbt}
\label{sec:spider}

The Spider 2.0 dbt benchmark scores an agent on realistic data transformation workflows against held out expected outputs. AutoFyn was given the goal of building and improving an agent for this task, and it constructed and optimized that agent autonomously across its rounds, using the benchmark's per case breakdown as the round signal. The resulting system, submitted as the SignalPilot Agent, holds the top position on the public leaderboard, and \autoref{fig:spider} places its score against the next four entries.

\subsection{Security Auditing}
\label{sec:security}

Here the verifier is the live exploit reproduction of \autoref{sec:verification}, and a finding counts only once it fires against the target build. To date AutoFyn has found over $150$ individual vulnerabilities across thirteen projects, among them Next.js, MetaMask, pnpm, Warp, LiteLLM, Langflow, Open WebUI, and RAGFlow. Related findings are bundled into a single advisory where they share a root cause, and we have submitted $43$ advisories from them, of which $16$ are confirmed by the maintainers of Next.js, MetaMask, pnpm, Warp, LiteLLM, Langflow, and Open WebUI, with the remaining $27$ still open. Each confirmed advisory carries an identifier a reader can check, and the full per project record is maintained in the project repository.

\autoref{fig:security-rounds} plots five of these runs round by round.

\section{Conclusion}
\label{sec:conclusion}
We described AutoFyn, an agent harness that bounds each round's context to a distilled state and admits a round as progress only on an external verifier's signal. It is interpreted as expert iteration over a contextual policy and the loop adapts a frozen model by changing what constructs its context rather than what sets its weights.

We ran the same loop ran in three domains differing only by the available verifiers and found impressive results. We make AutoFyn available as infrastructure for long horizon agents, with an auditable trail of every run.

\bibliographystyle{plainnat}
\bibliography{references}

\end{document}